\documentclass[10p,twocolumn,letterpaper]{IEEEtran}

\usepackage{graphicx}
\usepackage{amsmath}
\usepackage{xspace} 
\usepackage{balance}
\usepackage{url}
\usepackage{multirow}
\usepackage{fancyhdr}
\usepackage[pagebackref=false,breaklinks=true,colorlinks,bookmarks=false,urlcolor=blue,linkcolor=blue,citecolor=blue,pdftitle={Improved Foreground Detection via Block-based Classifier Cascade with Probabilistic Decision Integration},pdfauthor={Vikas Reddy, Conrad Sanderson, Brian C. Lovell}]{hyperref}

\def\Vec#1{{\mathbf{#1}}}
\def\Mat#1{{\mathbf{#1}}} 

\newcommand{\ie}{{ie.}\@\xspace}
\newcommand{\eg}{{eg.}\@\xspace}
\newcommand{\etal}{{et~al.}\@\xspace}

\begin{document}

\pagestyle{empty}

\title
  {
  \LARGE\bf
  Improved Foreground Detection via Block-based Classifier Cascade with Probabilistic Decision Integration%
  \thanks
    {
    \hrule\vspace{1ex}
    \noindent
    {\bf Published~in:}
    {IEEE Trans.~Circuits and Systems for Video Technology, Vol.~23, No.~1, 2013, pp.~83--93.}
    \href{http://dx.doi.org/10.1109/TCSVT.2012.2203199}{\bf dx.doi.org/10.1109/TCSVT.2012.2203199}
    Associated source code available at \href{http://arma.sourceforge.net/foreground/}{\bf http://arma.sourceforge.net/foreground/}
    Copyright \copyright~2013 IEEE. Personal use of this material is permitted.
    However, permission to use this material for any other purposes must
    be obtained from the IEEE by sending an email to
    \mbox{pubs-permissions@ieee.org}
    }
  }

\author
  {
  \large
  ~\\
  {Vikas Reddy, Conrad Sanderson, Brian C. Lovell}\\
  ~\\
  NICTA, PO Box 6020, St Lucia, QLD 4067, Australia\\
  University of Queensland, School of ITEE, QLD 4072, Australia\\
  Queensland University of Technology (QUT), Brisbane, QLD 4000, Australia
  }

\maketitle
\thispagestyle{empty}

\begin{abstract}

Background subtraction is a fundamental low-level processing task in numerous computer vision applications. 
The vast majority of algorithms process images on a pixel-by-pixel basis,
where an independent decision is made for each pixel.
A general limitation of such processing is that rich contextual information is not taken into account.
We propose a block-based method capable of dealing with noise, illumination variations and dynamic backgrounds,
while still obtaining smooth contours of foreground objects.
Specifically, image sequences are analysed on an overlapping block-by-block basis.
A low-dimensional texture descriptor obtained from each block is passed through an adaptive classifier cascade,
where each stage handles a distinct problem.
A probabilistic foreground mask generation approach then
exploits block overlaps to integrate interim block-level decisions
into final pixel-level foreground segmentation.
Unlike many pixel-based methods, ad-hoc post-processing of foreground masks is not required.
Experiments on the difficult Wallflower and I2R datasets 
show that the proposed approach obtains on average better results (both qualitatively and quantitatively)
than several prominent methods.
We furthermore propose the use of tracking performance as an unbiased approach for assessing
the practical usefulness of foreground segmentation methods,
and show that the proposed approach leads to considerable improvements
in tracking accuracy on the CAVIAR dataset.

\end{abstract}


\section{Introduction}
\label{sec:bg_introduction}

One of the fundamental and critical tasks in many computer-vision applications
is the segmentation of foreground objects of interest from an image sequence.
The accuracy of segmentation can significantly affect the overall performance of the application
employing it --- subsequent processing stages use only the foreground pixels rather than the entire frame.  
Segmentation is employed in diverse applications such as
tracking~\cite{porikli2003hbt, javed2006tracking},
action recognition~\cite{lv2007single},
gait recognition~\cite{wang2003silhouette},
anomaly detection~\cite{xiang2008incremental, reddy2011improved},
content based video coding~\cite{criminisi2006bilayer,vass1998automatic,neri1998automatic},
and
computational photography~\cite{agarwala2004interactive}. 

In the literature, foreground segmentation algorithms for an image sequence (video)
are typically based on segmentation via background modelling~\cite{piccardi2004bst,cvpr11brutzer},
which is also known as background subtraction~\cite{bouwmans2008background,bouwmans2010hbprcv}.
We note that foreground segmentation is also possible via optical flow analysis~\cite{barron1994performance},
an energy minimisation framework~\cite{criminisi2006bilayer, kohli2007dynamic}
as well as highly object specific approaches, such as detection of faces and pedestrians~\cite{viola2002robust,dalal2005histograms}.

Methods based on optical flow are prone to the aperture problem~\cite{trucco1998introductory}
and rely on movement --- stationary objects are not detected.
Methods based on energy minimisation require user intervention during their initialisation phase.
Specifically, regions belonging to foreground and background need to be explicitly labelled in order to build prior models.
This requirement can impose severe restrictions in applications where multiple
foreground objects are entering/exiting the scene (\eg~in surveillance applications).

Detection of specific objects is influenced by the training data set which
should ideally be exhaustive and encompass all possible  
variations/poses of the object --- in practice, this is hard to achieve. 
Furthermore, the type of objects to be detected must be known {\it a~priori}. 
These constraints can make object specific approaches unfavourable in certain surveillance environments
--- typically outdoors, where objects of various classes can be
encountered, including pedestrians, cars, bikes, and abandoned baggage.

In this work\footnote{This paper is a revised and extended version of our earlier work~\cite{reddy2010adaptive}.}
we focus on the approach of foreground segmentation via background modelling,
which can be formulated as a binary classification problem.
Unlike the other approaches mentioned above,
no constraints are imposed on the nature, shape or behaviour of foreground objects appearing in the scene.
The general approach is as follows.
Using a training image sequence,
a reference model of the background is generated.
The training sequence preferably contains only the dynamics of the background
(\eg~swaying branches, ocean waves, illumination variations, cast shadows).
Incoming frames are then compared to the reference model
and pixels or regions that do not fit the model (\ie~outliers) are labelled as foreground.
Optionally, the reference model is updated with areas that are deemed to be the background in the processed frames.

In general, foreground areas are selected in one of two ways:
{\bf (i)}~pixel-by-pixel, where an independent decision is made for each pixel,
and
{\bf (ii)}~region-based, where a decision is made on an entire group of spatially close pixels.
Below we briefly overview several notable papers in both categories.
As an in-depth review of existing literature is beyond the scope of this paper,
we refer the reader to several recent surveys for more 
details~\cite{piccardi2004bst,cvpr11brutzer,bouwmans2008background,bouwmans2010hbprcv,cristani2010background}.

The vast majority of the algorithms described in the literature belong to the pixel-by-pixel category.
Notable examples include techniques based on modelling the distribution of pixel values at each location.
For example,
Stauffer and Grimson~\cite{stauffer1999abm} model each pixel location by a Gaussian mixture model (GMM).
Extensions and improvements to this method include
model update procedures~\cite{kaewtrakulpong2001iab},
adaptively changing the
number of Gaussians per pixel~\cite{zivkovic2004iag}
and selectively filtering out pixels arising due to
noise and illumination, prior to applying GMM~\cite{teixeira2007object}. 
Some techniques employ non-parametric modelling --- 
for instance, Gaussian kernel density estimation~\cite{elgammal2000npm}
and a Bayes decision rule for classification~\cite{li2003foreground}.
The latter method models stationary regions of the image by colour features and
dynamic regions by colour co-occurrence features. 
The features are modelled by histograms.

Other approaches employ more complex strategies in order to improve 
segmentation quality in the presence of illumination variations and dynamic backgrounds. 
For instance, Han and Davis~\cite{6104064SVM} represent each pixel location by colour, gradient and Haar-like features. 
They use kernel density approximation to model features and a support vector machine for classification.
Parag~\etal~\cite{parag2006framework} automatically select a subset of features at each pixel location using a boosting algorithm.
Online discriminative learning~\cite{5604690Cheng} is also employed for real-time background subtraction using a graphics accelerator.
To address long and short term illumination changes separately,
\cite{5279449Shimada,Tanaka2009} maintain two distinct background models for colour and texture features.
Hierarchical approaches~\cite{5739511Guo, javed2008automated}
analyse data from various viewpoints (such as frame, region and pixel levels).
A related strategy which employs frame-level analysis to model 
background is subspace learning~\cite{oliver2000bcv, 4379761}.  
Although these methods process data at various levels, the classification is
still made at pixel level.

More recently, L{\'o}pez-Rubio~\etal~\cite{cviuLopezRubioB11} maintain a dual
mixture model at each pixel location for modelling the background and foreground distributions, respectively. 
The background pixels are modelled by a Gaussian distribution
while the foreground pixels are modelled by an uniform distribution.
The models are updated using a stochastic approximation technique.
Probabilistic self-organising maps have also been examined to model the
background~\cite{maddalena2008self, ijnsLopezRubioBD11}. 
To mitigate pixel-level noise,~\cite{ijnsLopezRubioBD11} 
also considers a given pixel's 8-connected neighbours prior
to its classification.

Notwithstanding the numerous improvements,
an inherent limitation of pixel--by--pixel processing is that rich contextual information is not taken into account.
For example, pixel-based segmentation algorithms may require ad-hoc post-processing
(\eg~morphological operations~\cite{Gonzalez_2007}) 
to deal with incorrectly classified and scattered pixels in the foreground mask. 

In comparison to the pixel-by-pixel category,
relatively little research has been done in the region-based category.
In the latter school of thought, each frame is typically split into blocks (or patches)
and the classification is made at the block-level
(\ie~effectively taking into account contextual information).
As adjacently located blocks are typically used,
a general limitation of region-based methods is that the generated foreground masks exhibit `blockiness' artefacts
(\ie rough foreground object contours).

Differences between blocks from a frame and the background can be measured by,
for example,
edge histograms~\cite{mason2001using}
and normalised vector distances~\cite{matsuyama2006background}.
Both of the above methods handle the problem of varying illumination but do not address dynamic backgrounds.
In methods~\cite{grabner2007time, lee2011hierarchical} for each block of the background,
a set of identical classifiers are trained using online boosting.
Blocks yielding a low confidence score are treated as foreground.
Other techniques within this family include
exploiting spatial co-occurrences of variations (\eg waving trees, illumination changes) 
across neighbouring blocks~\cite{seki2003bsb}, as well as 
decomposing a given video into spatiotemporal blocks
to obtain a joint representation of texture and motion
patterns~\cite{chan2008modeling, pokrajac2003spatiotemporal}.
The use of temporal analysis in the latter approach
aids in building good representative models but at an increased
computational cost.

In this paper we propose a robust foreground segmentation algorithm that belongs to the region-based category,
but is able to make the final decisions at the pixel level.
Briefly, a given image is split into overlapping blocks.
Rather than relying on a single classifier for each block,
an adaptive classifier cascade is used for initial labelling.
Each stage analyses a given block from a unique perspective.
The initial labels are then integrated at the pixel level.
A pixel is probabilistically classified as foreground/background
based on how many blocks containing that particular pixel have been classified as foreground/background.

The performance of foreground segmentation is typically evaluated
by comparing generated foreground masks with the corresponding ground-truth.
As foreground segmentation can be used in conjunction with tracking algorithms
(either as an aid or a necessary component~\cite{Sanin_PR_2012}),
we furthermore propose the use of object tracking performance as an additional
method for assessing the practical usefulness of foreground segmentation methods.

We continue the paper as follows.
In~Section~\ref{sec:bgs_proposed_algorithm} the proposed algorithm is described in detail.
Performance evaluation and comparisons with five other algorithms are given in
Section~\ref{sec:experiments}. The main findings and possible future directions are summarised in Section~\ref{sec:summary}.

\section{Proposed Foreground Detection Technique}
\label{sec:bgs_proposed_algorithm}

\noindent
The proposed technique has four main components:

\begin{itemize}

\item
Division of a given image into overlapping blocks,
followed by generating a low-dimensional descriptor for each block.

\item
Classification of each block into foreground or background,
where each block is processed by a cascade comprised of three classifiers.

\item
Model reinitialisation to address scenarios where a sudden and significant scene change
can make the current background model inaccurate.

\item
Probabilistic generation of the foreground mask,
where the classification decisions for all blocks are integrated
into final pixel-level foreground segmentation.

\end{itemize}

\noindent
Each of the components is explained in more detail in the following sections.

\subsection{Blocking and Generation of Descriptors}
\label{subsec:Feature Extraction}

Each image is split into blocks which are considerably smaller than the size of
the image (\eg~2$\times$2, 4$\times$4, \ldots, 16$\times$16), 
with each block overlapping its neighbours by a configurable amount of
pixels (\eg~1, 2, \ldots, 8) in both the horizontal and vertical directions.
Block overlapping can also be interpreted as block advancement. For instance,
maximum overlapping between blocks corresponds to block advancements by 1~pixel.

2D~Discrete Cosine Transform~(DCT) decomposition is employed to obtain a relatively 
robust and compact description of each
block~\cite{Gonzalez_2007}. 
Image noise and minor variations are effectively ignored by keeping only several low-order DCT coefficients
which reflect the average intensity and low frequency information~\cite{Sanderson_ICB_2009}.
Specifically, for a block located at {\small $(i,j)$},
four coefficients per colour channel are retained (based on preliminary experiments),
leading to a 12~dimensional descriptor:

\begin{small}
\begin{equation}
  \Vec{d}_{(i,j)}
  =
  \left[
    c^{[r]}_0, \cdots, c^{[r]}_3,~~
    c^{[g]}_0, \cdots, c^{[g]}_3,~~
    c^{[b]}_0, \cdots, c^{[b]}_3
  \right]^T
  \label{feature_vec}
\end{equation}%
\end{small}%

\noindent
where {\small $c^{[k]}_n$} denotes the {\small $n$}-th DCT coefficient
from the {\small $k$}-th colour channel,
with {\small $k \in \left\{ r, g, b \right\}$}.

\subsection{Classifier Cascade}
\label{subsec:multi-stage_Classifier}

Each block's descriptor is analysed sequentially by three classifiers,
with each classifier using location specific parameters.
As soon as one of the classifiers deems that the block is part of the background,
the remaining classifiers are not consulted.

The first classifier handles dynamic backgrounds (such as waving trees, water surfaces and fountains),
but fails when illumination variations exist.
The second classifier analyses if the anomalies in the descriptor are due to illumination variations.
The third classifier exploits temporal correlations (that naturally exists in image sequences)
to partially handle changes in environmental conditions and minimise spurious false positives.
The three classifiers are elucidated below.

\subsubsection{\bf Probability measurement}
\label{subsubsec:probability}

The first classifier employs a multivariate Gaussian model for each of the background blocks.
The likelihood of descriptor {\small $\Vec{d}_{(i,j)}$} belonging to the background class is found via:

\begin{small}
\begin{equation}
  p\left(\Vec{d}_{(i,j)} \right)
  =
  \frac
    {
    \exp
    \left\{
      -
      \frac{1}{2}
      \left[\Vec{d}_{(i,j)} - \Vec{\mu}_{(i,j)} \right]^T
      \Mat{\Sigma}^{-1}_{(i,j)}
      \left[\Vec{d}_{(i,j)} - \Vec{\mu}_{(i,j)} \right]
    \right\}
    }
    {
    \left( 2\pi \right)^\frac{D}{2}
    \left| \Mat{\Sigma}_{(i,j)}  \right|^{\frac{1}{2}}
    }
  \label{normal_dist}
\end{equation}
\end{small}

\noindent
where {\small $\Vec{\mu}_{(i,j)}$} and {\small $\Mat{\Sigma}_{(i,j)}$}
are the mean vector and
covariance matrix for location {\small $(i,j)$}, respectively,
while {\small $D$} is the dimensionality of the descriptors.
For ease of implementation and reduced computational load, the dimensions are
assumed to be independent and hence the covariance matrix is diagonal.

To obtain {\small $\Vec{\mu}_{(i,j)}$} and {\small $\Mat{\Sigma}_{(i,j)}$},
the first few seconds of the sequence are used for training.
To allow the training sequence to contain moving foreground objects,
a robust estimation strategy is employed instead of directly obtaining the parameters.
Specifically, for each block location a two-component Gaussian mixture model is trained,
followed by taking the absolute difference of the weights of the two Gaussians.
If the difference is greater than {\small $0.5$} (based on preliminary experiments),
we retain the Gaussian with the dominant weight.
The reasoning is that the less prominent Gaussian
is modelling moving foreground objects and/or other outliers.
If the difference is less than {\small $0.5$},
we assume that no foreground objects are present
and use all available data for that particular block location 
to estimate the parameters of the single Gaussian.
More involved approaches for dealing with foreground clutter during training are given in~\cite{Baltieri_2010,Reddy_IVP_2011}.

If~{\small $p(\Vec{d}_{(i,j)} ) \geq {T}_{(i,j)}$},
the corresponding block is classified as background.
The value of {\small ${T}_{(i,j)}$} is equal to {\small $p(\Vec{t}_{(i,j)} )$},
where   
{\small $\Vec{t}_{(i,j)} = \Vec{\mu}_{(i,j)} + 2 \operatorname{diag} (\Mat{\Sigma}_{(i,j)})^{\frac{1}{2}}$}.
Here the square root operation is applied element-wise.
Under the diagonal covariance matrix constraint,
this threshold covers about 95\% of the distribution~\cite{duda2001pattern}.

If a block has been classified as background,
the corresponding Gaussian model is updated
using the adaptation technique similar to Wren \etal~\cite{wren1997pfinder}.
Specifically, the mean and diagonal covariance vectors are updated as follows:

\begin{small}
\begin{align}
\Vec{\mu}_{(i,j)}^{new} = & ~ (1 - \rho) \Vec{\mu}_{(i,j)}^{old} + \rho \Vec{d}_{(i,j)} 
\label{eqn:update_param_mu}
\\
~ \nonumber \\
\Mat{\Sigma}_{(i,j)}^{new} = & ~
  (1 - \rho) \Mat{\Sigma}_{(i,j)}^{old} \nonumber \\
& +
  \rho (\Vec{d}_{(i,j)} - \Vec{\mu}_{(i,j)}^{new}) (\Vec{d}_{(i,j)} - \Vec{\mu}_{(i,j)}^{new})^T
\label{eqn:update_param_dcov}
\end{align}%
\end{small}%

\subsubsection{\bf Cosine distance}

The second classifier employs a distance metric based on the cosine of the angle subtended between two vectors.
Empirical observations suggest the angles subtended by descriptors
obtained from a block exposed to varying illumination are almost the same.
A similar phenomenon was also observed in RGB colour space~\cite{kim2005rtf}.

If block {\small $(i,j)$} has not been classified as part of the background by the previous classifier,
the cosine distance is computed using:

\begin{small}
\begin{equation}
  \operatorname{cosdist} ( \Vec{d}_{(i,j)}, \Vec{\mu}_{(i,j)} )
  =
  1 -
  \frac
  { \Vec{d}_{(i,j)}^T ~ \Vec{\mu}_{(i,j)} }
  { \parallel \Vec{d}_{(i,j)} \parallel  \parallel  \Vec{\mu}_{(i,j)} \parallel }
  \label{cosine_dist}
\end{equation}%
\end{small}%

\noindent
where {\small $\Vec{\mu}_{(i,j)}$} is from Eqn.~(\ref{normal_dist}).
If \mbox{\small $\operatorname{cosdist} (\Vec{d}_{(i,j)}, \Vec{\mu}_{(i,j)} )
\leq C_1$}, block {\small $(i,j)$} is deemed as background.
The value of $C_1$ is set to a low value such that it results in slightly
more false positives than false negatives. 
This ensures a low probability of misclassifying foreground objects as
background. 
However, the surplus false positives are eliminated during the creation of the
foreground mask (Section~\ref{subsec:Foreground Mask}).
Based on preliminary results, the constant $C_1$ is set to
0.1\% of the maximum value 
(for a cosine distance metric the maximum value is unity).

\subsubsection{\bf Temporal correlation check}

For each block, the third classifier takes into account the current descriptor
as well as the corresponding descriptor from the previous image,
denoted as {\small $\Vec{d}_{(i,j)}^{[\operatorname{prev}]}$}.
Block {\small $(i,j)$} is labelled as part of the background
if the following two conditions are satisfied:

\begin{enumerate}

\item
{\small $\Vec{d}_{(i,j)}^{[\operatorname{prev}]}$}  was classified as background;

\item
{\small $\operatorname{cosdist}
(\Vec{d}_{(i,j)}^{[\operatorname{prev}]},\Vec{d}_{(i,j)} ) \leq C_2$}.

\end{enumerate}

\noindent
Condition~1 ensures the cosine distance measured in Condition~2
is not with respect to a descriptor classified as foreground.
As the sample points are consecutive in time and should be almost identical
if $\Vec{d}_{(i,j)}$ belongs to background,
we use \mbox{\small $C_2 = 0.5 \times C_1$}.

\subsection{Model Reinitialisation}
\label{subsec:Model Reinitialisation}

A scene change might be too quick and/or too \mbox{severe} for the adaptation and classification strategies used above
(\eg~severe illumination change due to lights being switched on in a dark room).
As such, the existing background model can wrongly detect a very large portion of the image as foreground. 

Model reinitialisation is triggered if a `significant' portion of each image 
is consistently classified as foreground for a reasonable period of time.
Specifically, the criteria for defining significant portion is dependent on
parameters such as scene dynamics and size of foreground objects.
Based on preliminary evaluations, a threshold value of 70\% appears to work reasonably well. 
In order to ensure the model quickly adapts to the new environment, 
reinitialisation is invoked as soon as this phenomenon is consistently observed 
for a time period of at least {\small $\frac{1}{2}$} second 
(\ie~about 15 frames when sequences are captured at 30 fps).
The corresponding images are accumulated and are used to rebuild the statistics of the new scene.
Due to the small amount of retraining data, the covariance matrices are kept as is,
while the new means are obtained as per the estimation method described in Section~\ref{subsubsec:probability}.

\subsection{Probabilistic Foreground Mask Generation}
\label{subsec:Foreground Mask}

\begin{figure}[!t]
\centering
  \begin{minipage}{0.8\columnwidth}
  \centering
      \begin{minipage}{0.45\columnwidth}
      \centerline{\includegraphics[width=0.6\columnwidth]{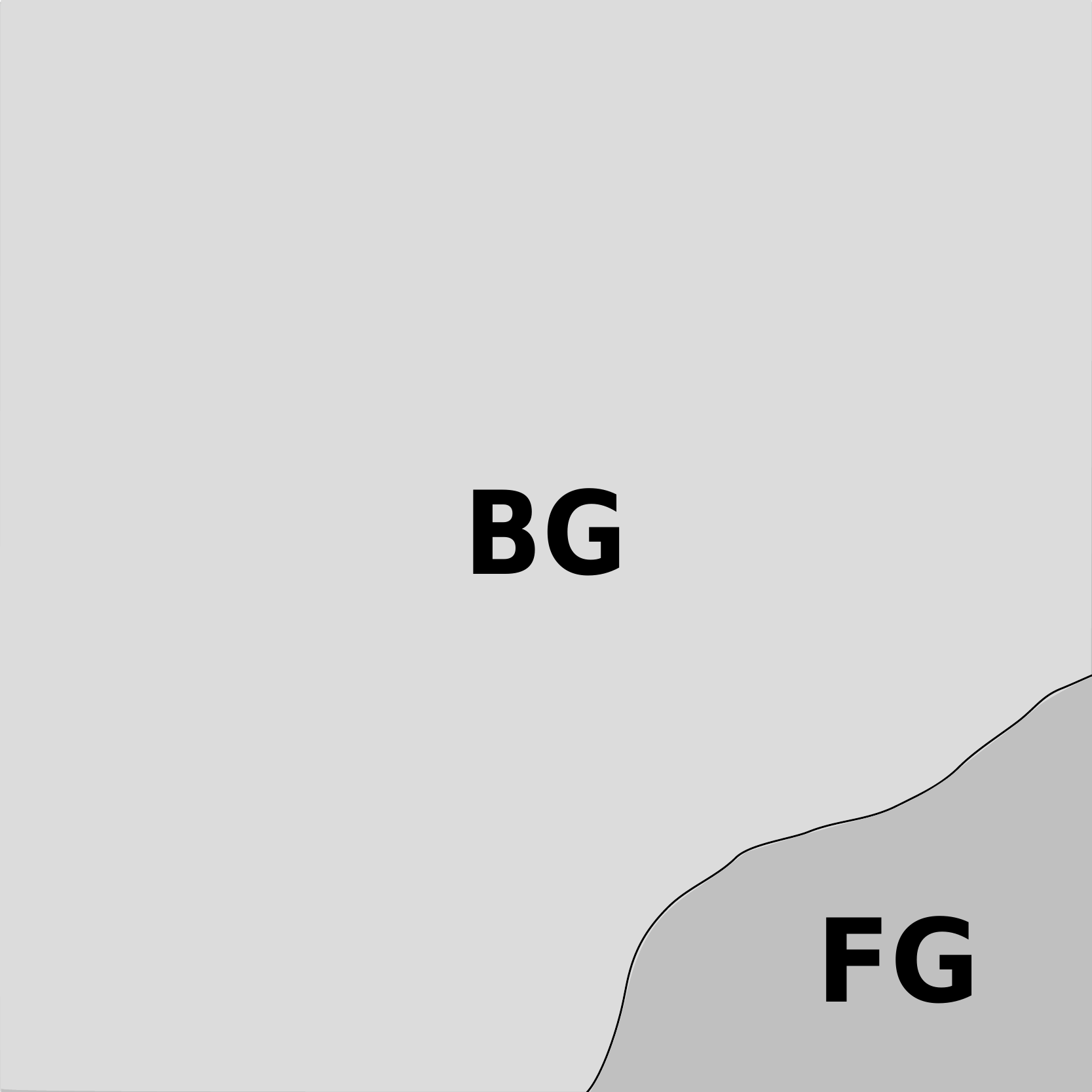}}
      \end{minipage}
      \hfill
      \begin{minipage}{0.45\columnwidth}
      \centerline{\includegraphics[width=0.6\columnwidth]{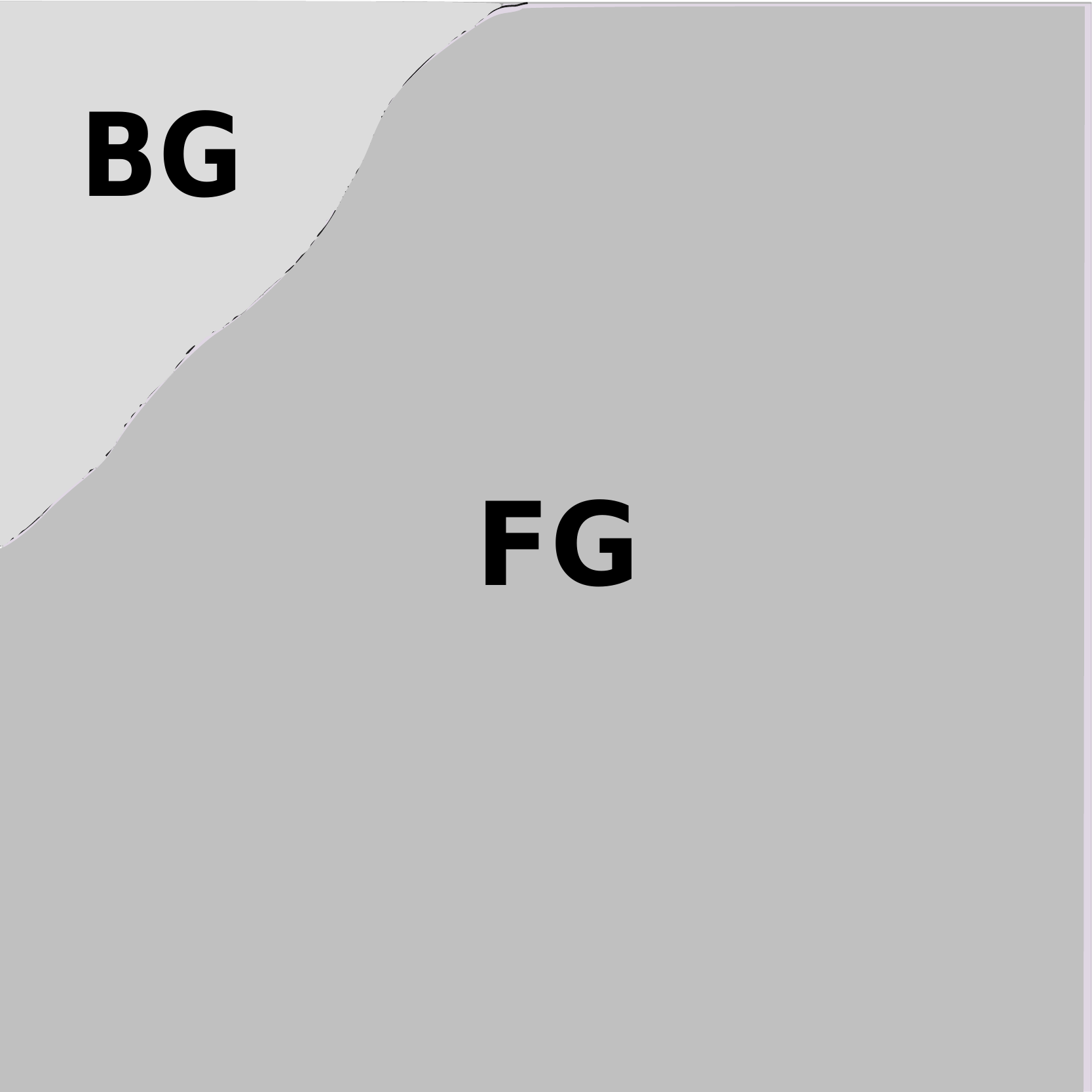}}
      \end{minipage}
  \end{minipage}  
  
  ~
  
   \begin{minipage}{0.8\columnwidth}
     \begin{minipage}{0.45\columnwidth}
       \centerline{\small Block A}
     \end{minipage}
     \hfill
     \begin{minipage}{0.45\columnwidth}
       \centerline{\small Block B}
     \end{minipage}
   \end{minipage}
    
  \caption
    {
    Without taking into account block overlapping,
    misclassification is inevitable at the pixel level
    whenever a given block has both foreground (FG) and background (BG) pixels.
    Classifying Block A as background results in a few false negatives (foreground pixels classified as background)
    while classifying Block B as foreground results in a few false positives (background pixels classified as foreground).
    }
\label{fig:foreground_background_blocks}
\end{figure}

In typical block based classification methods, misclassification is inevitable
whenever a given block has foreground and background pixels
(examples are illustrated in Fig.~\ref{fig:foreground_background_blocks}).
We exploit the overlapping nature of the block-based analysis to alleviate this inherent problem.
Each pixel is classified as foreground only if a significant proportion
of the blocks that contain that pixel are classified as foreground.
In other words, a pixel that was misclassified a few times prior to mask generation
can be classified correctly in the generated foreground mask. 
This decision strategy, similar to majority voting,
effectively minimises the number of errors in the output.
This approach is in contrast to conventional methods,
such as those based on Gaussian mixture models~\cite{kaewtrakulpong2001iab}, 
kernel density estimation~\cite{elgammal2000npm} and codebook models~\cite{kim2005rtf},
which do not have this built-in `self-correcting' mechanism.

Formally, let the pixel located at {\small $(x,y)$} in image {\small ${I}$} be denoted as {\small $I_{(x,y)}$}.
Furthermore, let {\small $B^{\operatorname{fg}}_{(x,y)}$} be the number of blocks
containing pixel {\small $(x,y)$} that were classified as foreground (fg),
and
{\small $B^{\operatorname{total}}_{(x,y)}$} be the total number of blocks containing pixel {\small $(x,y)$}.
We define the probability of foreground being present in {\small $I_{(x,y)}$} as:

\begin{small}
\begin{equation}
  P \left( \operatorname{fg} ~|~ {I_{(x,y)}}\right) = 
    B^{\operatorname{fg}}_{(x,y)} ~/~ B^{\operatorname{total}}_{(x,y)}
\end{equation}
\end{small}

\noindent
If {\small $P \left( \operatorname{fg} ~|~ {I_{(x,y)}}\right) \geq 0.90$}
(based on preliminary analysis),
pixel {\small $I_{(x,y)}$} is labelled as part of the foreground.

\section{Experiments}
\label{sec:experiments}

In this section, we first provide a brief description of the datasets used in
our experiments in Section~\ref{subsec:datasets}.
We then evaluate the effect of two 
key parameters (block size and  block advancement)
and the contribution of the three classifier stages to 
overall performance 
in Sections~\ref{subsec:effects_of_paramters} 
and~\ref{subsec:contribution_of_classifiers}, respectively.  

For comparative evaluation, we conducted two sets of
experiments: {\bf (i)}~subjective and objective evaluation of foreground segmentation efficacy,
 using datasets with available ground-truths;
{\bf (ii)}~comparison of the effect of the various foreground segmentation
methods on tracking performance.
The details of the experiments are described 
in Sections~\ref{subsec:Evaluation by ground-truth similarity} 
and~\ref{subsec:Evaluation by Tracking}, 
respectively. 

The proposed algorithm\footnote{\bf Source code for the proposed algorithm is available from \mbox{\url{http://arma.sourceforge.net/foreground/}}}
was implemented in C++ with the aid of
Armadillo~\cite{Armadillo_2010} and OpenCV libraries~\cite{Bradski2008}. 
All experiments were conducted on a standard 3~GHz machine.

\subsection{Datasets}
\label{subsec:datasets}

\begin{figure}[!tb]
  \centering
  \begin{minipage}{1\columnwidth}
    \includegraphics[width=0.48\textwidth]{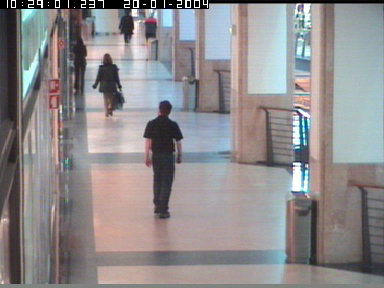}
    \hfill
    \includegraphics[width=0.48\textwidth]{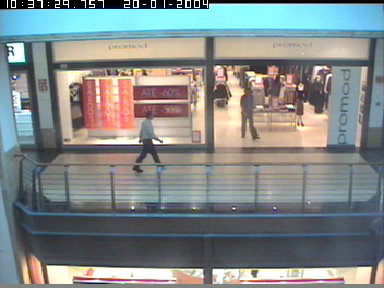}
   \end{minipage}
   \caption
     {
     Example frames from the second subset of the CAVIAR dataset,
     used for evaluating
     the influence of various foreground detection algorithms
     on tracking performance.
     } 
  \label{fig:CAVIAR_examples}
\end{figure}

We use three datasets for the experiments:
I2R\footnote{\url{http://perception.i2r.a-star.edu.sg/bk_model/bk_index.html}},
Wallflower\footnote{\url{http://research.microsoft.com/en-us/um/people/jckrumm/WallFlower/TestImages.htm}},
and
CAVIAR\footnote{\url{http://homepages.inf.ed.ac.uk/rbf/CAVIARDATA1/}}.
The I2R dataset has nine sequences captured in diverse and challenging environments 
characterised by complex backgrounds such as waving trees, fountains, 
and escalators. 
Furthermore, the dataset also exhibits the phenomena of illumination variations
and cast shadows. 
For each sequence there are 20 randomly selected images for which the ground-truth foreground masks are available.
The Wallflower dataset has seven~sequences,
with each sequence being a representative of a distinct problem encountered in
background modelling~\cite{toyama1999wpa}.
The background is subjected to various phenomena which include sudden and
gradual lighting changes, dynamic motion, camouflage, foreground aperture, 
bootstrapping and movement of background objects within the scene.
Each sequence has only one ground-truth foreground mask. 
The second subset of CAVIAR, used for the tracking experiments,
has 52 sequences with tracking ground truth data (\ie~object positions).
Example images from the three datasets are given in
Figures~\ref{fig:results_comp123}, \ref{fig:results_comp123_wf} and~\ref{fig:CAVIAR_examples}.

\subsection{Effects of Block Size and Advancement (Overlapping)}
\label{subsec:effects_of_paramters}

In this section we evaluate the effect of block size and block advancement to the overall performance.
For quantitative evaluation we adopted the F-measure metric used by Brutzer~\etal~\cite{cvpr11brutzer},
which quantifies how similar the obtained foreground mask is to the ground-truth:

\begin{equation}
  \operatorname{\it F-measure} = 2 \frac{recall \cdot precision}{recall + precision}
  \label{similarity_measure}
\end{equation}%

\noindent
where {$\operatorname{\it F-measure}$}~{$\in [0,1]$},
while {$precision$} and {$recall$} are given by $\frac{tp}{tp + fp}$  and
$\frac{tp}{tp + fn}$, respectively. 
The notations {$tp$}, {$fp$} and {$fn$}
are total number of true positives, false positives and false negatives
(in terms of pixels), respectively.
The higher the {$\operatorname{\it F-measure}$} value,
the more accurate the foreground segmentation.

\begin{table}[!tb]
  \centering
  \begin{small}
  \caption
    { 
    Accuracy of foreground estimation for various block sizes on the I2R and
    Wallflower datasets, with the block advancement fixed at 1 (\ie~maximum
    overlap). 
    Accuracy was measured by {$\operatorname{\it F-measure}$}
    averaged over all frames where ground-truth is available. 
    The `mean' column indicates the mean of the values obtained for the two datasets.
    }
  \label{tab:different_block_size_results}
    \begin{tabular}{c|c|c|c} \hline
      \multirow{2}{*}{Block Size} & \multicolumn{3}{c}{Average $\operatorname{\it F-measure}$}  \\ \cline{2-4}
                                     & ~~I2R~~  & Wallflower&  ~~mean~~ \\ \hline
      
{2$\times$2}    & 0.726     & 0.588       & 0.657 \\
{4$\times$4}    & \textbf{0.791}  & 0.633       & 0.712 \\  \hline
{6$\times$6}    & 0.790     & 0.714       & 0.752 \\
{8$\times$8}    & 0.780     & 0.733       & \textbf{0.756}  \\  \hline
{10$\times$10}    & 0.760       & \textbf{0.735}      & 0.735 \\
{12$\times$12}    & 0.732     & 0.729       & 0.731 \\  \hline
{14$\times$14}    & 0.704     & 0.715         & 0.710 \\
{16$\times$16}    & 0.659     & 0.692         & 0.675 \\  \hline

  \end{tabular}
  \end{small} 
\end{table}

Table~\ref{tab:different_block_size_results}
shows the performance of the proposed algorithm for block sizes ranging from 2$\times$2 to 16$\times$16,
with the block advancement fixed at 1 (\ie~maximum overlap between blocks).
The~optimal block size for the I2R dataset is 4$\times$4,
with the performance being quite stable from 4$\times$4 to 8$\times$8.
For the Wallflower dataset the optimal size is 10$\times$10,
with similar performance obtained using 8$\times$8 to 12$\times$12.
By taking the mean of the values obtained for each block size across both datasets,
the overall optimal size appears to be 8$\times$8.
This block size is used in all following experiments.

\begin{figure*}[!tb]

  \begin{minipage}{\textwidth}
  
    \begin{minipage}{0.15\textwidth}
      \centerline{\includegraphics[width=\columnwidth]{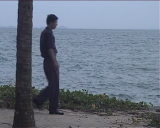}}
      \centerline{\footnotesize\bf(a)}
    \end{minipage}
    \hfill
    \begin{minipage}{0.15\textwidth}
      \centerline{\includegraphics[width=\columnwidth]{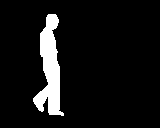}}
      \centerline{\footnotesize\bf(b)}
    \end{minipage}
    \hspace{0.1ex}
    \vline
    \vline
    \hspace{0.5ex}
    \begin{minipage}{0.15\textwidth}
      \centerline{\includegraphics[width=\columnwidth]{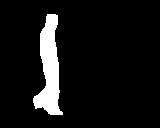}}
      \centerline{\footnotesize\bf(c)}
    \end{minipage}
    \hfill
    \begin{minipage}{0.15\textwidth}
      \centerline{\includegraphics[width=\columnwidth]{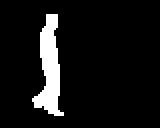}}
      \centerline{\footnotesize\bf(d)}
    \end{minipage}
    \hfill
    \begin{minipage}{0.15\textwidth}
      \centerline{\includegraphics[width=\columnwidth]{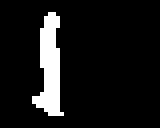}}
      \centerline{\footnotesize\bf(e)}
    \end{minipage}
    \hfill
    \begin{minipage}{0.15\textwidth}
      \centerline{\includegraphics[width=\columnwidth]{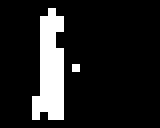}}
      \centerline{\footnotesize\bf(f)}
    \end{minipage}
  \end{minipage}
  \vspace{0.25ex}
  \caption
    {
    {\bf(a)} An example frame from the I2R dataset, {\bf(b)} its corresponding ground-truth foreground mask.
    Using the proposed method with a block size of $8\times8$,
    the foreground masks obtained for various degrees of block advancement:
    {\bf(c)} 1 pixel, {\bf(d)} 2 pixels, {\bf(e)} 4 pixels, and {\bf(f)} 8 pixels (\ie no overlap).
    }
  \label{fig:fg_mask_cmp_blk_ovp}
\end{figure*}

\begin{figure*}[!tb]
  \centering
  \vspace{2ex}
  \begin{minipage}{1.0\textwidth}
  \begin{minipage}{0.49\textwidth}
  \includegraphics[width = 1\textwidth]{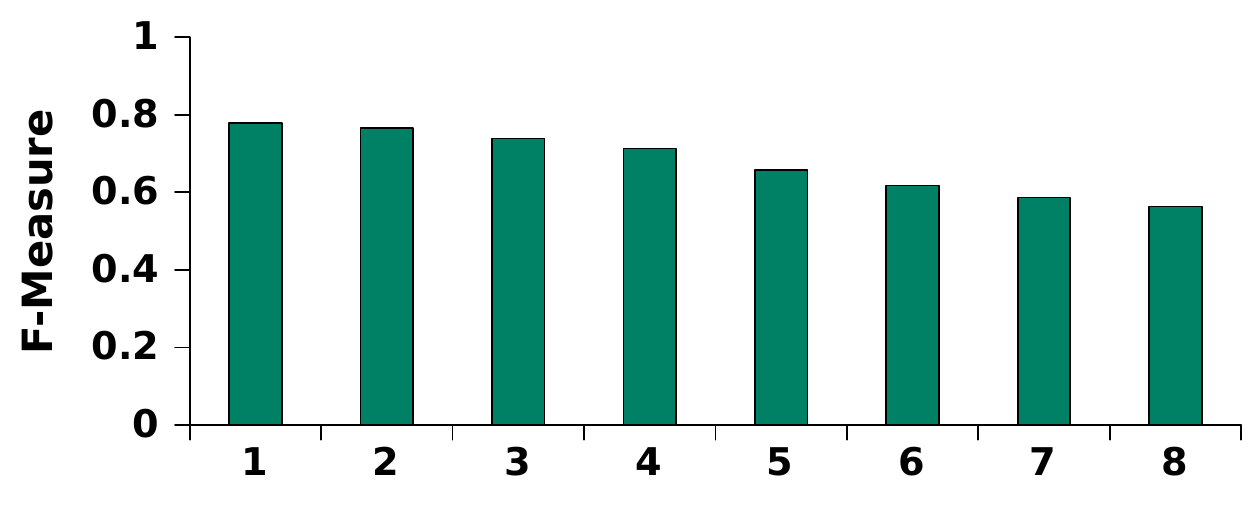}
  \end{minipage}
  \hfill
  \begin{minipage}{0.49\textwidth}
  \includegraphics[width=1\textwidth]{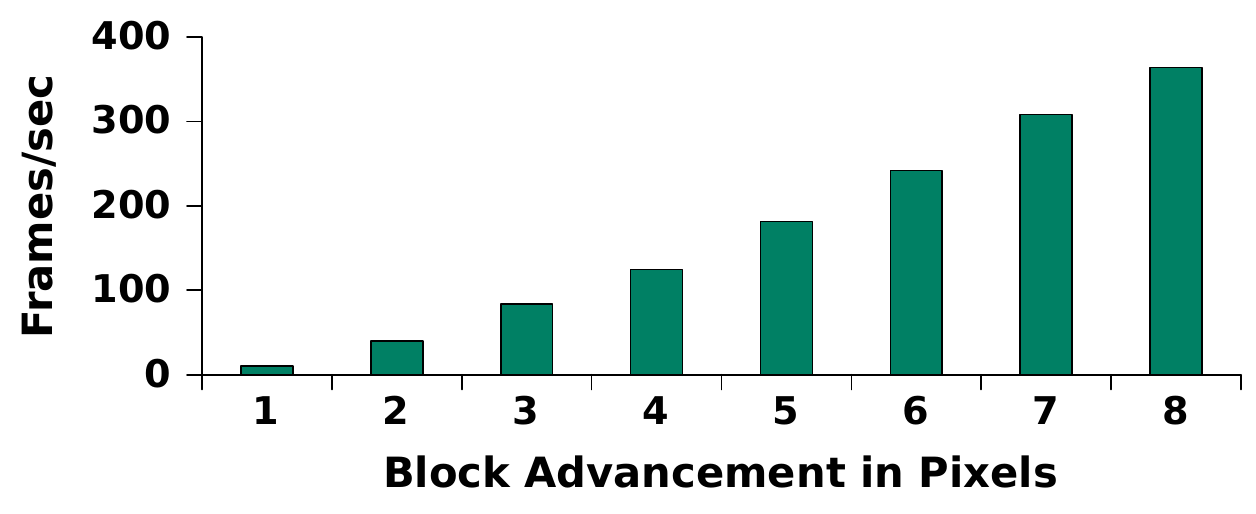}
  \end{minipage}
  \end{minipage}
  
  \vspace{2.0ex}
  
  \begin{minipage}{1.0\textwidth}
  \begin{minipage}{0.49\textwidth}
  \centerline{\bf{(a)}} 
  \end{minipage}
  \hfill
  \begin{minipage}{0.49\textwidth}  
  \centerline{\bf{(b)}}
  \end{minipage}
  \end{minipage}
   
  \caption
    { 
    Effect of block advancements on:
    {\bf (a)}~{$\operatorname{\it F-measure}$} value and
    {\bf (b)}~processing speed in terms of frames per second obtained using the I2R dataset. 
    A~considerable gain in processing speed is achieved as the advancement between blocks increases,
    at the expense of a gradual decrease in {$\operatorname{\it F-measure}$} values.
    }
  \label{fig:Similarity_vs_displacement}
\end{figure*}

Figures~\ref{fig:fg_mask_cmp_blk_ovp} and~\ref{fig:Similarity_vs_displacement}
show the effect of block advancement on foreground segmentation accuracy and processing speed on the I2R dataset. 
As the block size is fixed to 8$\times$8,
block advancement of 8 pixels (between successive blocks) indicates no overlapping,
while block advancement of 1 pixel denotes maximum overlap.
The smaller the block advancement (\ie~higher overlap),
the higher the accuracy and smoother object contours,
at the expense of a considerable increase in the computational load (due to more blocks that need to be processed).
A block advancement of 1 pixel achieves the best {$\operatorname{\it
F-measure}$} value of 0.78, at the cost of low processing speed (10 frames per second).
Increasing the block advancement to 2 pixels somewhat decreases the
{$\operatorname{\it F-measure}$} value to 0.76, but the processing speed raises to 40 frames per second.

\begin{figure*}[!tb]
  \begin{center}
  \vspace{3ex}
  \begin{minipage}{1\textwidth}
  \begin{minipage}{0.02\textwidth}
    \centerline{\footnotesize\bf(a)}
  \end{minipage}
  \begin{minipage}{0.95\textwidth}   
    \includegraphics[width=0.152\textwidth]{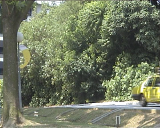}
    \hfill
    \includegraphics[width=0.152\textwidth]{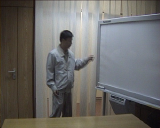}
    \hfill
    \includegraphics[width=0.152\textwidth]{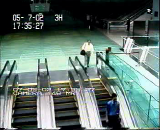}
    \hfill
    \includegraphics[width=0.152\textwidth]{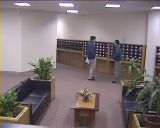}
    \hfill
    \includegraphics[width=0.152\textwidth]{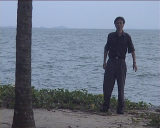}
    \hfill
    \includegraphics[width=0.152\textwidth]{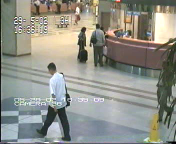}  
  \end{minipage}

  \vspace{1.5ex}
  \begin{minipage}{0.02\textwidth}
    \centerline{\footnotesize\bf(b)}
  \end{minipage}
  \begin{minipage}{0.95\textwidth}   
    \includegraphics[width=0.152\textwidth]{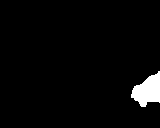}
    \hfill
    \includegraphics[width=0.152\textwidth]{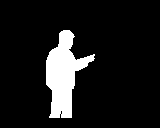}
    \hfill
    \includegraphics[width=0.152\textwidth]{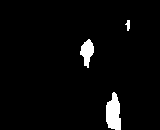}
    \hfill
    \includegraphics[width=0.152\textwidth]{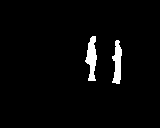}
    \hfill
    \includegraphics[width=0.152\textwidth]{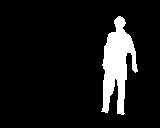}
    \hfill
    \includegraphics[width=0.152\textwidth]{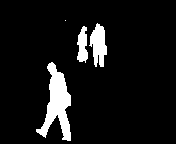}  
  \end{minipage}
   
   \vspace{1ex}
    
   \hrule
    
   \vspace{1ex}
   
  \begin{minipage}{0.02\textwidth}
    \centerline{\footnotesize\bf(c)}
  \end{minipage}
  \begin{minipage}{0.95\textwidth}   
    \includegraphics[width=0.152\textwidth]{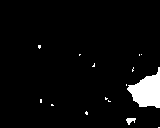}
    \hfill
    \includegraphics[width=0.152\textwidth]{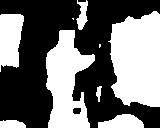}
    \hfill
    \includegraphics[width=0.152\textwidth]{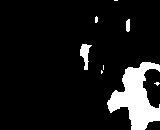}
    \hfill
    \includegraphics[width=0.152\textwidth]{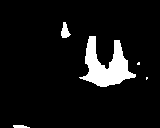}
    \hfill
    \includegraphics[width=0.152\textwidth]{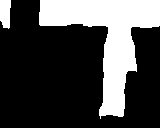}
    \hfill
    \includegraphics[width=0.152\textwidth]{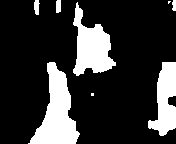}  
  \end{minipage}
   
  \vspace{1.5ex}
  \begin{minipage}{0.02\textwidth}
    \centerline{\footnotesize\bf(d)}
  \end{minipage}
  \begin{minipage}{0.95\textwidth}   
    \includegraphics[width=0.152\textwidth]{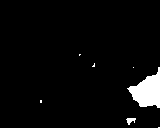}
    \hfill
    \includegraphics[width=0.152\textwidth]{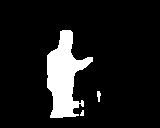}
    \hfill
    \includegraphics[width=0.152\textwidth]{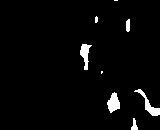}
    \hfill
    \includegraphics[width=0.152\textwidth]{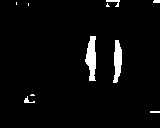}
    \hfill
    \includegraphics[width=0.152\textwidth]{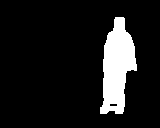}
    \hfill
    \includegraphics[width=0.152\textwidth]{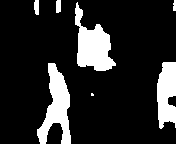}  
  \end{minipage}
  
  \vspace{1.5ex}
  \begin{minipage}{0.02\textwidth}
    \centerline{\footnotesize\bf(e)}
  \end{minipage}
   \begin{minipage}{0.95\textwidth}   
    \includegraphics[width=0.152\textwidth]{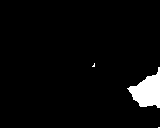}
    \hfill
    \includegraphics[width=0.152\textwidth]{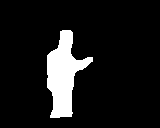}
    \hfill
    \includegraphics[width=0.152\textwidth]{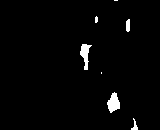}
    \hfill
    \includegraphics[width=0.152\textwidth]{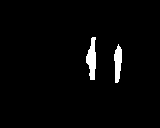}
    \hfill
    \includegraphics[width=0.152\textwidth]{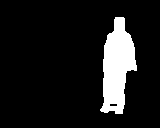}
    \hfill
    \includegraphics[width=0.152\textwidth]{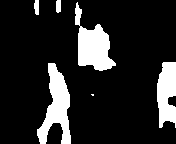}  
  \end{minipage}
 
  \end{minipage}
  \vspace{1ex}    
   \caption
     {
      {\bf (a)}~Example frames from I2R dataset.
      {\bf (b)}~Ground truth foreground masks.
      Foreground masks obtained by the proposed method using: 
      {\bf (c)}~the first classifier only, {\bf (d)}~combination of the first
      and second classifiers, {\bf (e)}~using all three classifiers. 
      Adding the second classifier considerably improves the segmentation quality,
      while the addition of the third classifier aids in minor reduction of false positives. 
      See Fig.~\ref{fig:bgs_classifiers_perf} for quantitative results.
      } 
  \label{fig:bgs_classifiers_qualitative_perf}
  \end{center}
\end{figure*}

\begin{figure*}[!tb]
\centering
  \includegraphics[height=0.7\columnwidth,width=1\columnwidth]{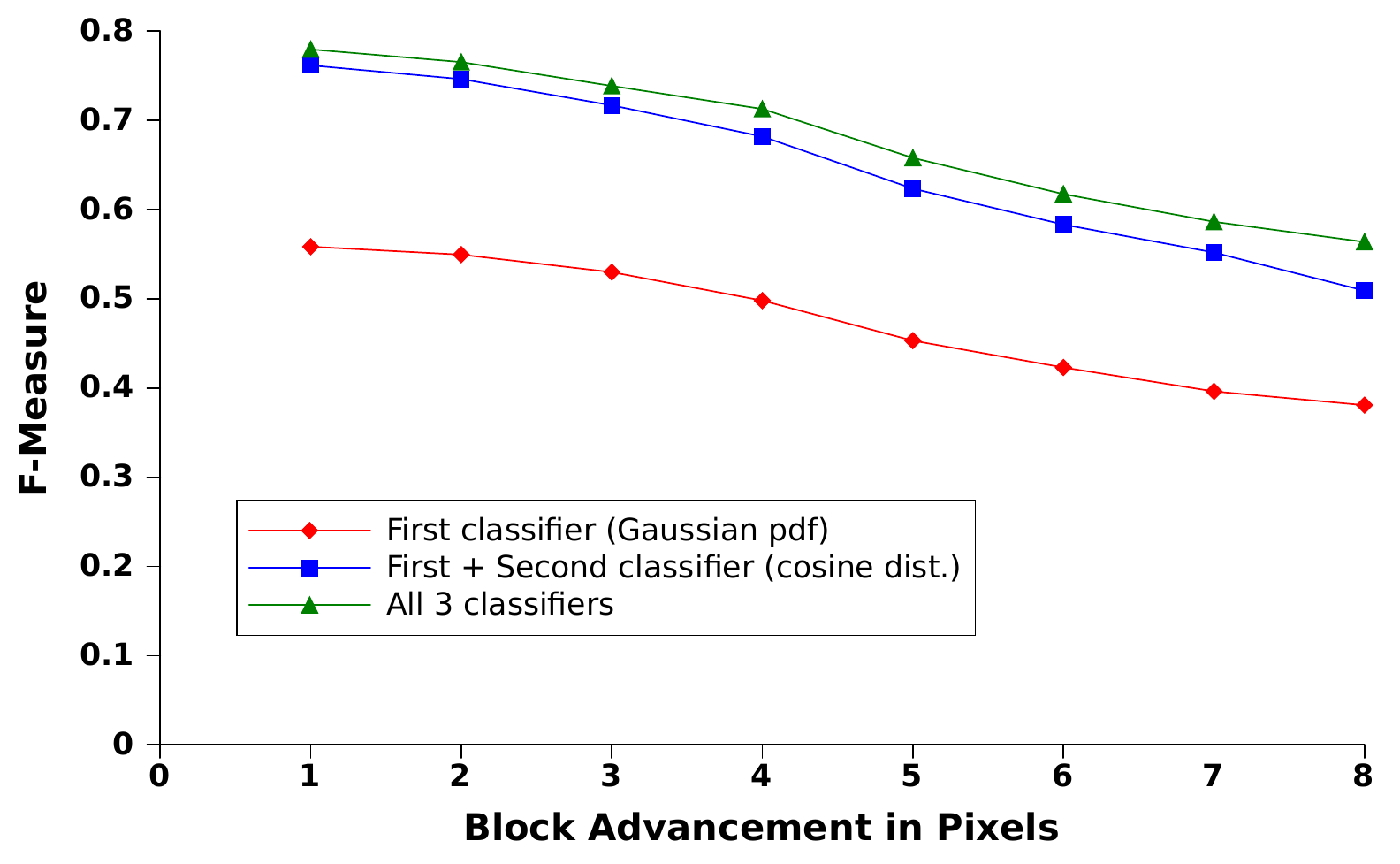}
  \caption
    {  
    Impact of individual classifiers on the overall segmentation quality for
    various block advancement values, using the I2R dataset.       
    The best results are achieved when all 3 classifiers are used (default
    configuration). 
    }
  \label{fig:bgs_classifiers_perf} 
\end{figure*}

\subsection{Contribution of Individual Classifier Stages}
\label{subsec:contribution_of_classifiers}

In the proposed algorithm, each classifier (see 
Section~\ref{subsec:multi-stage_Classifier}) handles a distinct problem such as
dynamic backgrounds and varying illuminations.
In this section, the influence of individual classifiers to the overall
segmentation performance is further investigated. 
We evaluate the segmentation 
quality using three separate configurations:
\textbf{(i)}~classification using the first classifier (based on multivariate Gaussian density function) alone,
\textbf{(ii)}~classification using a combination of the first classifier followed by the second (based on cosine distance),
\textbf{(iii)}~classification using all stages.
The the qualitative results of each configuration using the I2R dataset 
are shown in Fig.~\ref{fig:bgs_classifiers_qualitative_perf}.
The quantitative results of each configuration for various block advancements
are shown in Fig.~\ref{fig:bgs_classifiers_perf}.

We note that the best segmentation results are obtained for the default
configuration when all 3 classifiers are used. 
The next best configuration is the combination of the first and second
classifiers which independently inspect for scene changes occurring due 
to dynamic backgrounds and illumination variations, respectively.
The configuration comprising of only the first classifier yields the 
lowest {$\operatorname{\it F-measure}$} value, since background variations
due to illumination are not handled  effectively by it.

We note the impact of the third classifier appears to be minor
compared to that of the second, since it is aimed to
minimise the occasional false positives by  
examining the temporal correlations between consecutive frames
(see Section~\ref{subsec:multi-stage_Classifier}(c)).
The relative improvement in average {$\operatorname{\it F-measure}$} value
achieved by adding the second classifier is about 37\%, while adding the 
third gives further relative improvement of about~5\%.
Qualitative results of each configuration shown in Figure~\ref{fig:bgs_classifiers_qualitative_perf}
confirm the above observations.

\begin{figure*}[!t]
  \centering
  \includegraphics[width=1.0\textwidth]{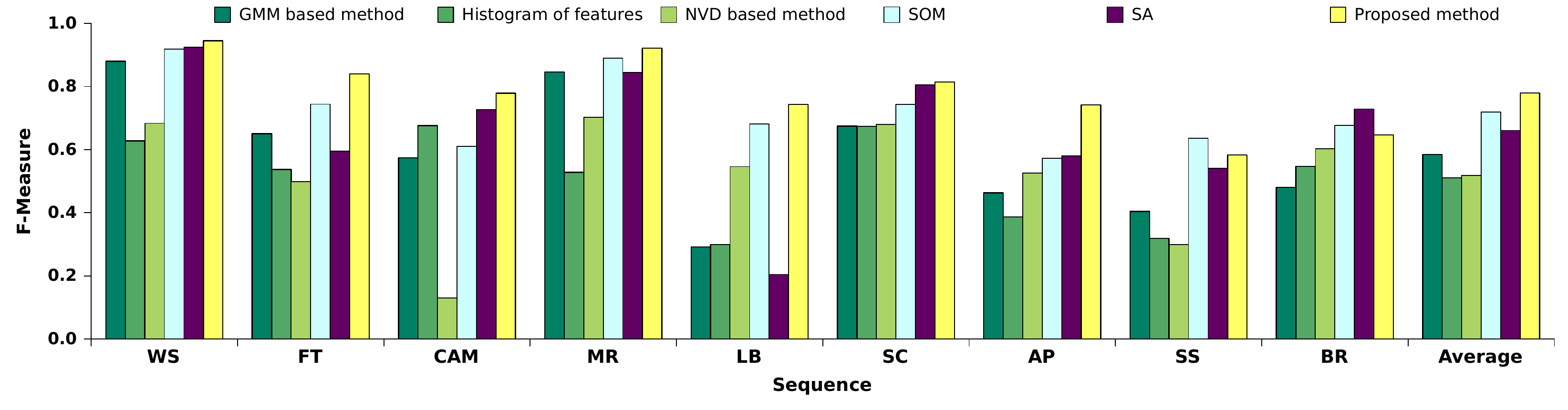}
  \caption
    {
    Comparison of {$\operatorname{\it F-measure}$} values
    (defined in Eqn.~(\ref{similarity_measure}))
    obtained on the I2R dataset using foreground segmentation methods based on
    GMMs~\cite{kaewtrakulpong2001iab},
    feature histograms~\cite{li2003foreground},
    NVD~\cite{matsuyama2006background},
    SOM~\cite{ijnsLopezRubioBD11},
    SA~\cite{cviuLopezRubioB11}
    and the proposed method.
    The higher the {$\operatorname{\it F-measure}$}
    (\ie~agreement with ground-truth), the~better the segmentation result.
    }
  \label{fig:Similarity_value_plot}

~

\end{figure*}

\begin{figure*}[!t]
  \centering
  \includegraphics[width=1.0\textwidth]{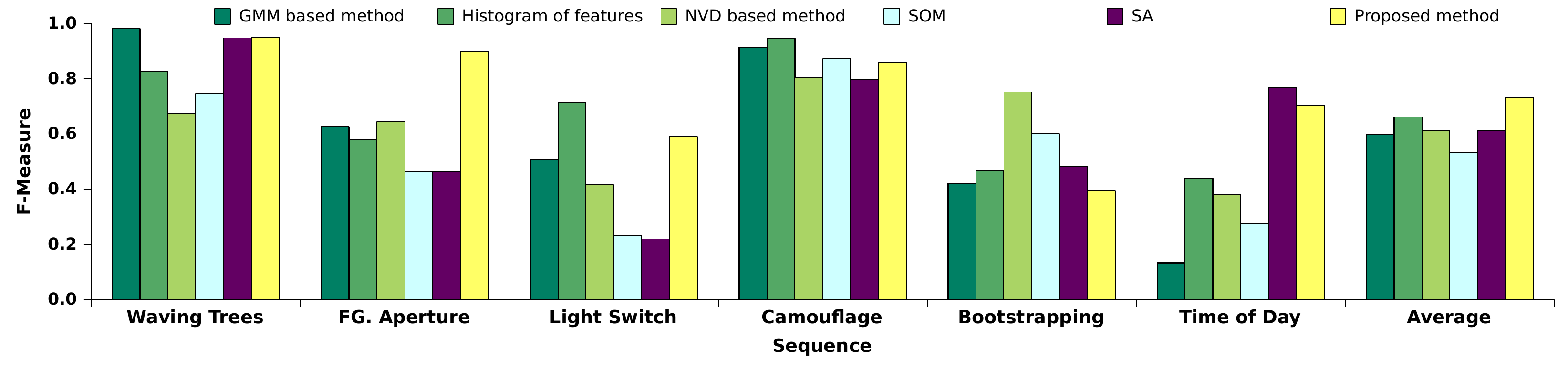}
  \caption
    {
    As per Fig.~\ref{fig:Similarity_value_plot}, but obtained on the Wallflower dataset.
    Due to the absence of true positives in the ground-truth for the \textit{moved object} sequence,
    the corresponding F-measure is zero for all algorithms.
    }
  \label{fig:Similarity_value_plot_wf}

~

\end{figure*}

\subsection{Comparative Evaluation by Ground-Truth F-measure}
\label{subsec:Evaluation by ground-truth similarity}
    
The proposed algorithm is compared with
segmentation methods based on
Gaussian mixture models (GMMs)~\cite{kaewtrakulpong2001iab},
feature histograms~\cite{li2003foreground}, probabilistic self
organising maps~(SOM)~\cite{ijnsLopezRubioBD11}, stochastic
approximation~(SA)~\cite{cviuLopezRubioB11} and normalised vector
distances~(NVD)~\cite{matsuyama2006background}. 
The first four methods classify individual pixels into foreground or background,
while the last method makes decisions on groups of pixels.

We used the OpenCV~v2.0~\cite{Bradski2008} implementations
for the GMM and feature histogram based methods with default parameters,
except for setting the learning parameter in GMM to 0.001.
Experiments showed that the above parameter settings produce
optimal segmentation performance.
We used the implementations made available by the authors' 
for SOM\footnote{\url{http://www.lcc.uma.es/~ezeqlr/fsom/fsom.html}}
and SA\footnote{\url{http://www.lcc.uma.es/~ezeqlr/backsa/backsa.html}} methods. 

Post-processing using morphological operations was required for the foreground 
masks obtained by the GMM, feature histogram and SOM methods,
in order to clean up the scattered error pixels.
For the GMM method, opening followed by closing using a 
{\small $3\times3$} kernel was performed,
while for the feature histogram method we enabled the \mbox{built-in} post-processor
(using default parameters suggested in the OpenCV implementation). 
We note that the proposed method does not require any such ad-hoc post-processing.

With the view of designing a pragmatic system, the same parameter settings were
used across all sequences (\ie~they were not optimised for any particular sequence).
Specifically, during deployment a practical system has to perform robustly in many scenarios.

\begin{figure}[!tb]
  \begin{minipage}{\columnwidth}
\begin{center}
  \begin{minipage}{0.04\columnwidth}
    \centerline{\footnotesize\bf(a)}
  \end{minipage}
  \begin{minipage}{0.94\columnwidth}   
    {\includegraphics[width=0.3\columnwidth, height=0.24\columnwidth]{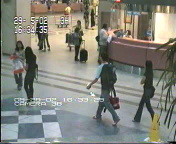}}
  \hfill
    {\includegraphics[width=0.3\columnwidth, height=0.24\columnwidth]{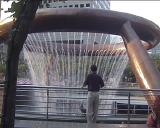}}
  \hfill
    {\includegraphics[width=0.3\columnwidth, height=0.24\columnwidth]{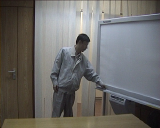}}
  \end{minipage}
\end{center}  
 \vspace{1ex}
\begin{center}
  \begin{minipage}{0.04\columnwidth}
    \centerline{\footnotesize\bf(b)}
  \end{minipage}
  \begin{minipage}{0.94\columnwidth}   
    {\includegraphics[width=0.3\columnwidth, height=0.24\columnwidth]{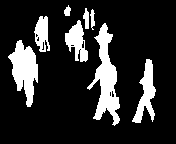}}
  \hfill
    {\includegraphics[width=0.3\columnwidth, height=0.24\columnwidth]{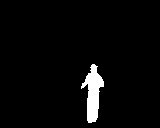}}
  \hfill
    {\includegraphics[width=0.3\columnwidth, height=0.24\columnwidth]{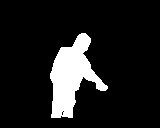}}
  \end{minipage}
\end{center} 
  \vspace{0.5ex}
   \hrule
   \vspace{0.5ex} 
 \begin{center}
  \begin{minipage}{0.04\columnwidth}
    \centerline{\footnotesize\bf(c)}
  \end{minipage}
  \begin{minipage}{0.94\columnwidth}   
    {\includegraphics[width=0.3\columnwidth, height=0.24\columnwidth]{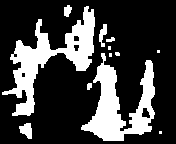}}
  \hfill
    {\includegraphics[width=0.3\columnwidth, height=0.24\columnwidth]{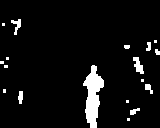}}
  \hfill
    {\includegraphics[width=0.3\columnwidth, height=0.24\columnwidth]{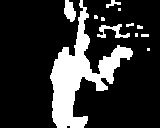}}
  \end{minipage}
\end{center}  
 \vspace{1ex}
\begin{center}
  \begin{minipage}{0.04\columnwidth}
    \centerline{\footnotesize\bf(d)}
  \end{minipage}
  \begin{minipage}{0.94\columnwidth}   
    {\includegraphics[width=0.3\columnwidth, height=0.24\columnwidth]{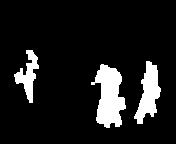}}
  \hfill
    {\includegraphics[width=0.3\columnwidth, height=0.24\columnwidth]{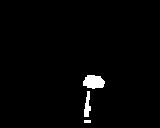}}
  \hfill
    {\includegraphics[width=0.3\columnwidth, height=0.24\columnwidth]{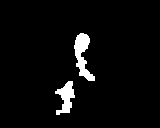}}
  \end{minipage}
\end{center}  
 \vspace{1ex}
\begin{center}
  \begin{minipage}{0.04\columnwidth}
    \centerline{\footnotesize\bf(e)}
  \end{minipage}
  \begin{minipage}{0.94\columnwidth}   
    {\includegraphics[width=0.3\columnwidth, height=0.24\columnwidth]{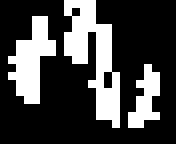}}
  \hfill
    {\includegraphics[width=0.3\columnwidth, height=0.24\columnwidth]{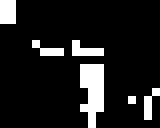}}
  \hfill
    {\includegraphics[width=0.3\columnwidth, height=0.24\columnwidth]{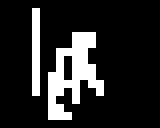}}
  \end{minipage}
\end{center}  
 \vspace{1ex}
\begin{center}
  \begin{minipage}{0.04\columnwidth}
    \centerline{\footnotesize\bf(f)}
  \end{minipage}
   \begin{minipage}{0.94\columnwidth}   
    {\includegraphics[width=0.3\columnwidth, height=0.24\columnwidth]{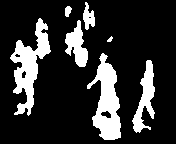}}
  \hfill
    {\includegraphics[width=0.3\columnwidth, height=0.24\columnwidth]{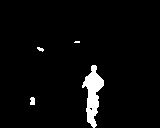}}
  \hfill
    {\includegraphics[width=0.3\columnwidth, height=0.24\columnwidth]{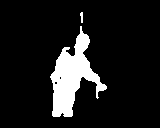}}
  \end{minipage}
\end{center}  
 \vspace{1ex}
\begin{center}
  \begin{minipage}{0.04\columnwidth}
    \centerline{\footnotesize\bf(g)}
  \end{minipage}
   \begin{minipage}{0.94\columnwidth}   
    {\includegraphics[width=0.3\columnwidth, height=0.24\columnwidth]{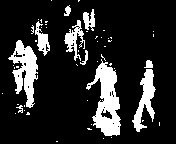}}
  \hfill
    {\includegraphics[width=0.3\columnwidth, height=0.24\columnwidth]{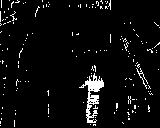}}
  \hfill
    {\includegraphics[width=0.3\columnwidth, height=0.24\columnwidth]{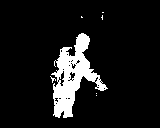}}
  \end{minipage}
\end{center}  
 \vspace{1ex}
 \begin{center}
  \begin{minipage}{0.04\columnwidth}
    \centerline{\footnotesize\bf(h)}
  \end{minipage}
  \begin{minipage}{0.94\columnwidth}   
    {\includegraphics[width=0.3\columnwidth, height=0.24\columnwidth]{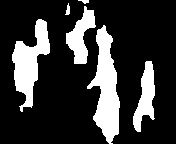}}
  \hfill
    {\includegraphics[width=0.3\columnwidth, height=0.24\columnwidth]{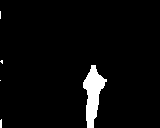}}
  \hfill
    {\includegraphics[width=0.3\columnwidth, height=0.24\columnwidth]{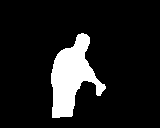}}
  \end{minipage}
\end{center}  
\vspace{1ex}
  \caption
    {
    {\bf (a)} Example frames from three video sequences in the I2R dataset. 
    {\it Left:}~people walking at an airport, with significant cast shadows.
    {\it Middle:}~people moving against a background of a fountain with varying
    illumination. {\it Right:}~a person walks in and out of a room where the
    window blinds are non-stationary, 
                  with illumination variations caused by automatic gain control 
                  of the camera.
    {\bf (b)} Ground-truth foreground mask, and foreground mask estimation using:
    {\bf (c)}~GMM based~\cite{kaewtrakulpong2001iab} 
             with morphological post-processing,~
    {\bf (d)}~feature~histograms~\cite{li2003foreground},
    {\bf (e)}~NVD~\cite{matsuyama2006background},~
    {\bf (f)}~SOM~\cite{ijnsLopezRubioBD11},
    {\bf (g)}~SA~\cite{cviuLopezRubioB11},
    {\bf (h)}~proposed~method.
    }
  \label{fig:results_comp123}

\end{minipage}
\end{figure}

\begin{figure}[!tb]
  \begin{minipage}{\columnwidth}

\begin{center}
  \begin{minipage}{0.04\columnwidth}
    \centerline{\footnotesize\bf(a)}
  \end{minipage}
  \begin{minipage}{0.94\columnwidth}   
    {\includegraphics[width=0.3\columnwidth, height=0.24\columnwidth]{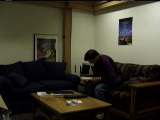}}
  \hfill
    {\includegraphics[width=0.3\columnwidth, height=0.24\columnwidth]{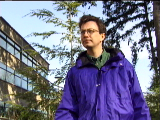}}
  \hfill
    {\includegraphics[width=0.3\columnwidth, height=0.24\columnwidth]{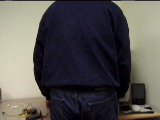}}
  \end{minipage}
\end{center}  
 \vspace{1ex}
\begin{center}
  \begin{minipage}{0.04\columnwidth}
    \centerline{\footnotesize\bf(b)}
  \end{minipage}
  \begin{minipage}{0.94\columnwidth}   
    {\includegraphics[width=0.3\columnwidth, height=0.24\columnwidth]{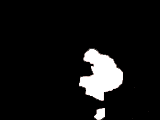}}
  \hfill
    {\includegraphics[width=0.3\columnwidth, height=0.24\columnwidth]{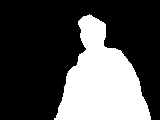}}
  \hfill
    {\includegraphics[width=0.3\columnwidth, height=0.24\columnwidth]{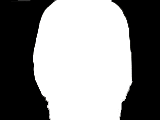}}
  \end{minipage}
\end{center}  
   \vspace{0.5ex}
   \hrule
   \vspace{0.5ex} 
\begin{center}
  \begin{minipage}{0.04\columnwidth}
    \centerline{\footnotesize\bf(c)}
  \end{minipage}
  \begin{minipage}{0.94\columnwidth}   
    {\includegraphics[width=0.3\columnwidth, height=0.24\columnwidth]{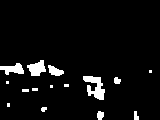}}
  \hfill
    {\includegraphics[width=0.3\columnwidth, height=0.24\columnwidth]{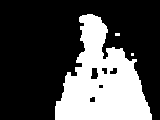}}
  \hfill
    {\includegraphics[width=0.3\columnwidth, height=0.24\columnwidth]{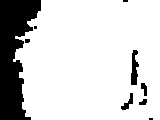}}
  \end{minipage}
\end{center}  
 \vspace{1ex}
\begin{center}
  \begin{minipage}{0.04\columnwidth}
    \centerline{\footnotesize\bf(d)}
  \end{minipage}
  \begin{minipage}{0.94\columnwidth}   
    {\includegraphics[width=0.3\columnwidth, height=0.24\columnwidth]{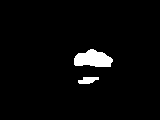}}
  \hfill
    {\includegraphics[width=0.3\columnwidth, height=0.24\columnwidth]{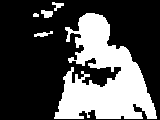}}
  \hfill
    {\includegraphics[width=0.3\columnwidth, height=0.24\columnwidth]{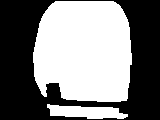}}
  \end{minipage}
\end{center}  
 \vspace{1ex}
\begin{center}
  \begin{minipage}{0.04\columnwidth}
    \centerline{\footnotesize\bf(e)}
  \end{minipage}
 \begin{minipage}{0.94\columnwidth}   
    {\includegraphics[width=0.3\columnwidth, height=0.24\columnwidth]{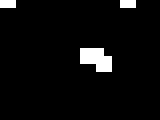}}
  \hfill
    {\includegraphics[width=0.3\columnwidth, height=0.24\columnwidth]{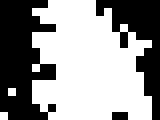}}
  \hfill
    {\includegraphics[width=0.3\columnwidth, height=0.24\columnwidth]{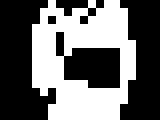}}
  \end{minipage}
\end{center}  
 \vspace{1ex}
  \begin{center}
  \begin{minipage}{0.04\columnwidth}
    \centerline{\footnotesize\bf(f)}
  \end{minipage}
  \begin{minipage}{0.94\columnwidth}   
    {\includegraphics[width=0.3\columnwidth, height=0.24\columnwidth]{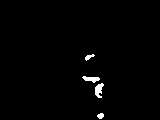}}
  \hfill
    {\includegraphics[width=0.3\columnwidth, height=0.24\columnwidth]{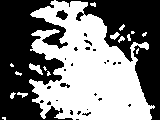}}
  \hfill
    {\includegraphics[width=0.3\columnwidth, height=0.24\columnwidth]{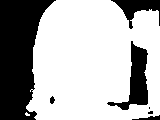}}
  \end{minipage}
  \end{center}
  \vspace{1ex}
   \begin{center}
  \begin{minipage}{0.04\columnwidth}
    \centerline{\footnotesize\bf(g)}
  \end{minipage}
  \begin{minipage}{0.94\columnwidth}   
    {\includegraphics[width=0.3\columnwidth, height=0.24\columnwidth]{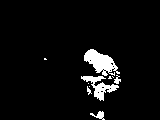}}
  \hfill
    {\includegraphics[width=0.3\columnwidth, height=0.24\columnwidth]{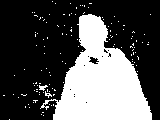}}
  \hfill
    {\includegraphics[width=0.3\columnwidth, height=0.24\columnwidth]{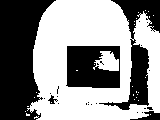}}
  \end{minipage}
\end{center}  
 \vspace{1ex}
\begin{center}
  \begin{minipage}{0.04\columnwidth}
    \centerline{\footnotesize\bf(h)}
  \end{minipage}
  \begin{minipage}{0.94\columnwidth}   
    {\includegraphics[width=0.3\columnwidth, height=0.24\columnwidth]{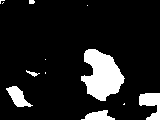}}
  \hfill
    {\includegraphics[width=0.3\columnwidth, height=0.24\columnwidth]{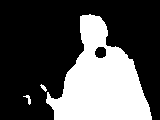}}
  \hfill
    {\includegraphics[width=0.3\columnwidth, height=0.24\columnwidth]{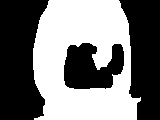}}
  \end{minipage}
\end{center}  
 \vspace{1ex}
 \end{minipage}
  \caption
    {
    As per Fig.~\ref{fig:results_comp123}, but using the Wallflower dataset.  
    {\it Left:}~room illumination gradually increases over time and a 
           person walks in and sits on the couch.
    {\it Middle:}~person walking against a background of strongly waving trees and the sky.
    {\it Right:}~a monitor displaying a~blue screen with rolling bars is
    occluded by a~person wearing blue coloured clothing.
    } 
  \label{fig:results_comp123_wf}
  
  
  ~
  
  ~
  
  ~
  
  ~
  
\end{figure}

We present both qualitative and quantitative analysis of the results.
Figs.~\ref{fig:Similarity_value_plot} and~\ref{fig:Similarity_value_plot_wf}
show quantitative results for the I2R and Wallflower datasets, respectively.
The corresponding qualitative results for three sequences from each dataset
are shown in Figs.~\ref{fig:results_comp123} and~\ref{fig:results_comp123_wf}.

In Fig.~\ref{fig:results_comp123},
the AP sequence (left column) has significant cast shadows of people moving
at an airport. The FT sequence (middle column) contains people moving against a
background of a fountain with varying illumination.
The MR sequence (right column) shows a person entering and leaving a room where 
the window blinds are non-stationary
and there are significant illumination variations caused by the automatic
 gain control of the camera.

In Fig.~\ref{fig:results_comp123_wf},
the {\it time of day} sequence (left column)
has a gradual increase in the room's illumination intensity over time.
A person walks in and sits on the couch. 
The {\it waving trees} sequence (middle column)
has a person walking against a background consisting of the sky and strongly waving trees.
In the \textit{camouflage} sequence (right column),
a monitor has a blue screen with rolling bars.
A person in blue coloured clothing walks in and occludes the monitor.

We note that output of the GMM based method (column~{\bf c} 
in Figs.~\ref{fig:results_comp123} and~\ref{fig:results_comp123_wf})
is sensitive to reflections, illumination changes and cast shadows.
While the histogram based method (column~{\bf d})
overcomes these limitations, it has a lot of false negatives.
The NVD based method (column~{\bf e}) is largely robust to illumination changes,
but fails to handle dynamic backgrounds and produces `blocky' foreground masks.
The SOM and SA based methods have relatively few false positives and negatives.
The results obtained by the proposed method (column~{\bf f})
are qualitatively better than those obtained by the other five methods,
having low false positives and false negatives.
However, we note that due to the the block-based nature of the analysis,
objects very close to each other tend to merge.

The quantitative results (using the F-measure metric)
obtained on the I2R and Wallflower datasets,
shown in Figs.~\ref{fig:Similarity_value_plot} and~\ref{fig:Similarity_value_plot_wf},
respectively, largely confirm the visual results.
On the I2R dataset the proposed method outperforms the other methods in most cases.
The next best method (SOM) obtained an average {$\operatorname{\it F-measure}$} value of 0.72,
while the proposed method achieved 0.78, representing an improvement of about~8\%. 

On the Wallflower dataset
the proposed method achieved considerably better results
for the {\it foreground aperture} sequence.
While for the remainder of the sequences the performance was roughly on par 
with the other methods,
the proposed method nevertheless still achieved the highest average 
{$\operatorname{\it F-measure}$} value.
The next best method (histogram of features)
obtained an average value of 0.66,
while the proposed method obtained 0.73,
representing an improvement of about~11\%.

We note that the performance of the proposed method on 
{\it Bootstrapping} sequence is lower. 
We conjecture that this is due to foreground objects
occluding background during the training phase. 
Robust background initialisation techniques~\cite{Baltieri_2010, Reddy_IVP_2011} 
capable of estimating the background in cluttered sequences could be used to
alleviate this problem.

\subsection{Comparative Evaluation by Tracking Precision \& Accuracy}
\label{subsec:Evaluation by Tracking}

We conducted a second set of experiments to evaluate
the performance of the segmentation methods in more pragmatic terms
rather than limiting ourselves to the traditional ground-truth evaluation approach.
To this effect, we evaluated the influence of the various foreground detection
algorithms on tracking performance. 
The foreground masks obtained from the detectors for each frame of the sequence
were passed as input to an object tracking system.
We have used a particle filter based tracker%
\footnote
  {
  Additional simulations with other tracking algorithms,
  such as blob matching, mean shift and mean shift with foreground feedback,
  yielded similar results.
  }
as implemented in the video surveillance module of OpenCV v2.0~\cite{Bradski2008}.
Here the foreground masks are used prior to tracking for initialisation purposes.

Tracking performance was measured with the two metrics proposed 
by Bernardin and Stiefelhagen~\cite{BernardinEtAl2008},
namely multiple object tracking precision (MOTP)
and multiple object tracking accuracy (MOTA).

Briefly, MOTP measures the average pixel distance between ground-truth locations of objects 
and their locations according to a tracking algorithm.
Ground truth objects and hypotheses are matched using Munkres' algorithm~\cite{munkres1957algorithms}.
MOTP is defined as:

\begin{small}
\begin{equation}
  \operatorname{MOTP} =  {\sum\nolimits_{i,t}d_{t}^{i}} ~ / ~ {\sum\nolimits_{t}c_{t}} 
\end{equation}
\end{small}

\noindent
where $d_{t}^{i}$ is the distance between object~$i$ and its corresponding hypothesis,
while $c_t$ is the number of matches found at time~$t$.
The lower the MOTP, the better.
 
MOTA accounts for object configuration errors, false positives, misses as well as mismatches.
It is defined as:

\begin{small}
\begin{equation}
   \operatorname{MOTA} =  1 - \frac {{\sum_{t}(m_{t} + fp_{t} + mme_{t})}}{\sum_{t}g_{t}}
\label{eqn:mota1}
\end{equation}
\end{small}
 
It measures accuracy in terms of the number of false negatives ($m$),
false positives ($fp$) and mismatch errors ($mme$) 
with respect to the number of ground truth objects ($g$). 
The higher the value, the better the accuracy. 
The MOTA value can become negative in certain circumstances
when the false negatives, false positives and mismatch errors are considerably large,
making the ratio in Eqn.~(\ref{eqn:mota1}) greater than unity~\cite{BernardinEtAl2008}.

\begin{table}[!tb]
  \centering
  \caption
    {
    Effect of various settings of block advancement on
    multiple object tracking accuracy (MOTA) in terms of percentage,
    and 
    multiple object tracking precision (MOTP) in terms of pixels. 
    Results are obtained on second subset of CAVIAR by using a particle-filter
    based tracking algorithm.
    }
  \label{tab:mota_motp_pxladv}
  \begin{tabular}{c|c|c} \hline
    \multirow{3}{*}{Block Advancement} & \multicolumn{2}{c}{\bf Tracking Metrics} \\ \cline{2-3}
                                       & {MOTA}             & {MOTP}              \\
                                       & (higher is better) & (lower is better)   \\ \hline
     1  & {\bf ~ 30.3}       & {\bf 11.7}  \\  
     2  & {~ 20.4}       & 11.8    \\ 
     4  & {~~ 8.6}       & 12.4    \\  
     8  & {$-$67.3}      & 14.7    \\ \hline
  \end{tabular}
\end{table}

\begin{table}[!tb]
  \centering
  \caption
    {
    As per Table.~\ref{tab:mota_motp_pxladv},
    but obtained by employing various foreground detection methods.
    }
  \label{tab:mota_motp_baseline}
  \begin{tabular}{c|c|c} \hline
    \multirow{3}{*}{Foreground detection} & \multicolumn{2}{c}{\bf Tracking Metrics} \\ \cline{2-3}
                                          & {MOTA}             & {MOTP}              \\
                                          & (higher is better) & (lower is better)   \\ \hline
     GMM based method~\cite{kaewtrakulpong2001iab}   & {~~ 27.2}       & {13.6}
     \\ NVD based method~\cite{matsuyama2006background} & {$-$24.9}      & 15.2    \\ 
     Histogram of features~\cite{li2003foreground}   & {~~ 13.7}       & 14.7    \\  
     SOM~\cite{ijnsLopezRubioBD11}                   & {~~ 26}         & 13.3    \\
     SA~\cite{cviuLopezRubioB11}             & {~~ 27.3}       & 13.0    \\
     Proposed method                 & {\bf ~~ 30.3}   & {\bf 11.7}  \\ \hline

  \end{tabular}

\end{table}

The performance result is the average performance of the 52 test sequences belonging to the second subset of CAVIAR.
To keep the evaluations more realistic,
the first few frames (200~frames) of each sequence are used to train  
the background model irrespective of the presence of 
foreground objects (\ie~background frames were not handpicked for training).

We first evaluated the tracking performance for various block advancements.
Results presented in Table~\ref{tab:mota_motp_pxladv} indicate that
a block advancement of 1~pixel obtains the best tracking performance,
while larger advancements lead to a decrease in performance.

Comparisons with GMM, histogram of features and NVD,
presented in Table~\ref{tab:mota_motp_baseline},
indicate that the proposed method leads to considerably better tracking performance.
For tracking accuracy (MOTA),
the next best method (SA) led to an average accuracy of 27.3\%,
while the proposed method led to 30\%.
For tracking precision (MOTP), 
the next best method (SA)
led to an average pixel distance of~13,
while the proposed method reduced the distance to~11.7.

\section{Main Findings}
\label{sec:summary}

Pixel-based processing approaches to foreground detection can be susceptible to noise,
illumination variations and dynamic backgrounds,
partly due to not taking into account rich contextual information.
In contrast, region-based approaches mitigate the effect of above phenomena 
but suffer from `blockiness' artefacts.
The proposed foreground detection method belongs to region-based category,
but at the same time is able segment smooth contours of foreground objects.

Contextual spatial information is employed through analysing each frame on an overlapping block-by-block basis.
The low-dimensional texture descriptor for each block alleviates the effect of image noise.
The model initialisation strategy allows the training sequence to contain moving foreground objects.
The adaptive classifier cascade analyses the descriptor from various perspectives before classifying the corresponding block as foreground.
Specifically, it checks if disparities are due to background motion or illumination variations,
followed by a temporal correlation check to minimise the occasional false positives emanating
due to background characteristics which were not handled by the preceding classifiers.

The probabilistic foreground mask generation approach integrates
the block-level classification decisions by exploiting the overlapping nature of the analysis,
ensuring smooth contours of the foreground objects 
as well as effectively minimising the number of errors.
Unlike many pixel-based methods, ad-hoc post-processing of foreground masks is not required.

Experiments conducted to evaluate the standalone performance 
(using the difficult Wallflower and I2R datasets)
show the proposed method obtains on average better results (both qualitatively and quantitatively)
than methods based on GMMs,
feature histograms, 
normalised vector distances,
self organising maps and
stochastic approximation.

We furthermore proposed the use of tracking performance as an unbiased approach for assessing
the practical usefulness of foreground segmentation methods,
and demonstrated that the proposed method leads to considerable improvements
in object tracking accuracy on the CAVIAR dataset.

\section*{Acknowledgements}

We thank Andres Sanin for helping with the tracking framework.
We also thank Sandra Mau, Mehrtash Harandi and the anonymous reviewers for their valuable feedback.
NICTA is funded by the Australian Government
as represented by the {\it Department of Broadband, Communications and the Digital Economy},
as well as the Australian Research Council through the {\it ICT Centre of Excellence} program.

\balance

\end{document}